\documentclass[utf8]{FrontiersinHarvard} 
\usepackage{url,hyperref,lineno,microtype,subcaption}
\usepackage[onehalfspacing]{setspace}
\usepackage[dvipsnames]{xcolor}
\usepackage{url}
\usepackage{graphicx}
\graphicspath{{./images/}}
\usepackage{caption,subcaption}

\usepackage{xspace}
\makeatletter
\DeclareRobustCommand\onedot{\futurelet\@let@token\@onedot}
\def\@onedot{\ifx\@let@token.\else.\null\fi\xspace}

\def\eg{\textit{e.g}\onedot} 
\def\ie{\textit{i.e}\onedot} 
 
\def\etc{\textit{etc}\onedot}

\makeatother

\nolinenumbers
\def\keyFont{\fontsize{8}{11}\helveticabold }
\def\firstAuthorLast{Avogaro {et~al.}}
\def\Authors{Andrea Avogaro\,$^{1}$, Federico Cunico\,$^{1}$, Bodo Rosenhahn\,$^{2}$ and Francesco Setti\,$^{3,*}$}
\def\Address{
$^{1}$Department of Computer Science, University of Verona, Verona, Italy\\ 
$^{2}$Institute for Information Processing, Leibniz University Hannover, Hannover, Germany\\
$^{3}$Department of Engineering for Innovation Medicine, University of Verona, Verona, Italy
}

\def\corrAuthor{Francesco Setti}

\def\corrEmail{francesco.setti@univr.it}

\begin{document}
\onecolumn
\firstpage{1}
\title[Markerless HPE for Biomedical Applications: A Survey]{Markerless Human Pose Estimation for Biomedical Applications: A Survey}
\author[\firstAuthorLast ]{\Authors} 
\address{} 
\correspondance{} 
\extraAuth{}
\maketitle
\begin{abstract}
Markerless Human Pose Estimation (HPE) proved its potential to support decision making and assessment in many fields of application. HPE is often preferred to traditional marker-based Motion Capture systems due to the ease of setup, portability, and affordable cost of the technology. 
However, the exploitation of HPE in biomedical applications is still under investigation.
This review aims to provide an overview of current biomedical applications of HPE. In this paper we examine the main features of HPE approaches and discuss whether or not those features are of interest to biomedical applications. We also identify those areas where HPE is already in use and present peculiarities and trends followed by researchers and practitioners.
We include here 25 approaches to HPE and more than 40 studies of HPE applied to motor development assessment, neuromuscolar rehabilitation and gait \& posture analysis.
We conclude that markerless HPE offers great potential for extending diagnosis and rehabilitation outside hospitals and clinics, towards the paradigm of \emph{remote medical care}. 
\tiny
\keyFont{ \section{Keywords:} markerless motion capture, human pose estimation, biomedical applications, general movement analysis, gait analysis, neuromuscular, rehabilitation, surbey.}
\end{abstract}

\section{Introduction}
\label{sec:intro}

Movement data contain a huge amount of meaningful information about our body and brain status; so much that a dedicated research field named \emph{movement science} was born.
Humans are interested in analyzing their own or others' movement for the most varied reasons: humans track and forecast neighbors' movements for non-verbal communication as well as to avoid collisions in collaborative environments, a sport coach oversees athlete's movements to optimize performances and prevent injuries, and a physiotherapist analyzes patient's mobility to monitor the effectiveness of rehabilitation after a stroke or severe injury.
Beyond medical applications, the understanding and reconstruction of human motion opened several commercial fields in the domains of games and movies, intelligent production, collaborative robotics, autonomous driving, home security (fall detections) and many more.
Humans learn to move and to interpret others movements over many years of development from childhood on, so that it is becoming somehow ``natural'' for all of us. For this reason, translating this knowledge to machines is a complex task, that requires first accurately capturing human motion and subsequently making decisions or interpretations on top of it.

The interest in computer vision based capturing human motion dates back some centuries~\citep{bregler1,MoeslundHMSurvey,mundermann2006evolution}.
At this time, the core challenges have been in the hardware (capturing, storing and transmitting synchronized multi-view images), the calibration of cameras, extraction of features, multi-view matching and 3D reconstruction. A more historic overview is given in ~\citep{Klette2008}.

The continuous commercial developments, \eg driven by the games and movie industry, heavily boosted human pose estimation for biomedical applications.

Thus, MoCap and HPE can be seen as different facets of the same field of research. Nowadays, a standard setting is to use computer vision, machine learning, and deep learning techniques to identify key landmarks on the body (\eg knees, elbows, \etc) from visual inputs in the form of images/videos of RGB channels, depth, or a combination of the two. This input can be used for skeletal-based reconstruction of human motion and subsequent tasks.

Motion capture systems have been used in biomedical applications for a few decades, allowing the quantification and identification of complex movement patterns and impairments~\citep{Richards99,scott2022healthcare}.
Biomechanical information can be used to inform clinical decision-making, like the diagnosis of neurological diseases or assessing rehabilitation programs~\citep{fogel2018artificial}.
Systems used in clinical evaluations are predominantly 3D marker-based MoCap setups. 
These systems can provide accurate and reliable information to clinicians, and there are proven benefits in patients' outcomes as a result of better informed decisions. Significant demonstrable benefits are decreased number of unnecessary surgeries and more efficient diagnosis and treatments, both in terms of time and costs.
Nevertheless, benefits come with high costs due to the complex infrastructure required (\ie infrared cameras, powerful computers, \etc) and the need for technically experienced laboratory staff.
These aspects de-facto prevent the adoption of these technologies on a vast scale, also producing barriers to the accessibility to such devices.
Moreover, these systems are not widely used across multiple health conditions and tend to be focused on a few specific patient populations, like Parkinson's disease and Cerebral Palsy~\citep{salami2019long}.

HPE, also referred to as markerless MoCap, may offer several advantages over marker-based solutions, mostly related to a simpler setup.
Indeed, markerless HPE do not require to position reflective markers on the patient, resulting in reduced effort and more efficiency for clinicians and being less invasive on the patients.
Hardware setups are often cheaper since input data can be acquired with standard RGB or depth sensors rather than IR cameras.
These data can be easily acquired with common household devices (\eg webcams, smartphones, videogames consoles) which offers the exciting potential for deploying such systems with minimal costs and effort~\citep{stenum2021applications}.
The drawback is a lower accuracy and robustness that often require to find a trade-off between usability and performances~\citep{kanko2021concurrent}.

In a recent study the authors observe that current HPE and tracking algorithms do not fit the needs of movement science due to the poor performances in estimating variables like joints velocity, acceleration, and forces which are crucial for movement science~\citep{seethapathi2019movement}.
While this can be true for some specific problems, we argue that many real-world applications do not actually need such high precision and they do actually benefit from state-of-the-art HPE. Indeed, there are nowadays several approaches that can provide reliable skeletal reconstructions~\citep{cao2017openpose,tu2020voxelpose}, ready to be adopted in real-world applications.
Alternatively, HPE can be used as an assistive tool for clinicians rather than an automatic machine, allowing medical personnel to take advantage of automatic pose estimation, correct wrong estimates, and interpret quantitative data to make decisions. This approach can save lots of time since the alternative is to manually annotate videos, which is a tedious and error-prone operation.

A number of review papers appeared in the last years on this topic, but previous works either limit their analysis to specific approaches like 3D pose estimation~\citep{desmarais2021review}, single camera~\citep{scott2022healthcare} and deep learning models~\citep{cronin2021using}, or they focus on specific clinical applications like motor development assessment~\citep{leo2022video,silva2021future}, neuromuscolar rehabilitation~\citep{arac2020machine,cherry2023opportunities} and gait analysis~\citep{wade2022applications}.

In this paper, we focus on the biomedical applications of HPE. We examine the main features of HPE approaches, discussing whether or not those features are of interest to biomedical applications. Furthermore, this work will provide an overview of state-of-the-art methods and analyze their suitability for the medical context (Sec.~\ref{sec:hpe}). We will also identify applications in the biomedical field where such technologies are already studied or tested, presenting some peculiarities of these applications and the trends followed by researchers and practitioners (Sec.~\ref{sec:applications}). We will also briefly present a few commercial products for HPE (Sec.~\ref{sec:commercial}). Finally, we will discuss the limitations of current approaches and opportunities for new research on this topic (Sec.~\ref{sec:discussion}).

\section{Human Pose Estimation}
\label{sec:hpe}

Human Pose Estimation concerns the ability of a machine to identify the position of the joints of a human body in 3D space or in an image \citep{deeppose}. 

Several commercial systems have been developed to solve this problem with estimation errors up to less than a millimeter, but there are still a number of limitations. 

The currently established gold standard
 requires positioning of optical markers, \ie spheres of reflective material close to the joints of the subject (following a predefined protocol for marker placement) and the illumination of the scene with ad-hoc light sources. This assumes a predefined recording volume and collaboration of the subject. Both is not always possible in healthcare applications (\eg remote care or working with infants). The presence of marker can also affect the performances both physically and psychologically. Moreover, systems like Vicon\footnote{\url{https://www.vicon.com}} and Optitrack\footnote{\url{https://www.optitrack.com}} are very expensive, making them not affordable for many applications.
This explains the demand for HPE that can work with common devices like standard cameras or smartphones. 

In the last decades many solutions have been proposed, each one with specific pros and cons. Please note, that this survey is limited to computer vision and deep learning based approaches with a core focus on biomedical applications. Thus, we have to omit impressive works from animation, games or surveilance which make use of different sensors, special prior knowledge on the motions or multi-modal setups.
In the following we will discuss the main features of HPE methods for biomedical applications and provide a taxonomy of these approaches which is summarized in Table~\ref{tab:summary}.

\subsection{2D vs. 3D estimation}

\begin{figure}[t]
    \centering
    \includegraphics[height=6cm]{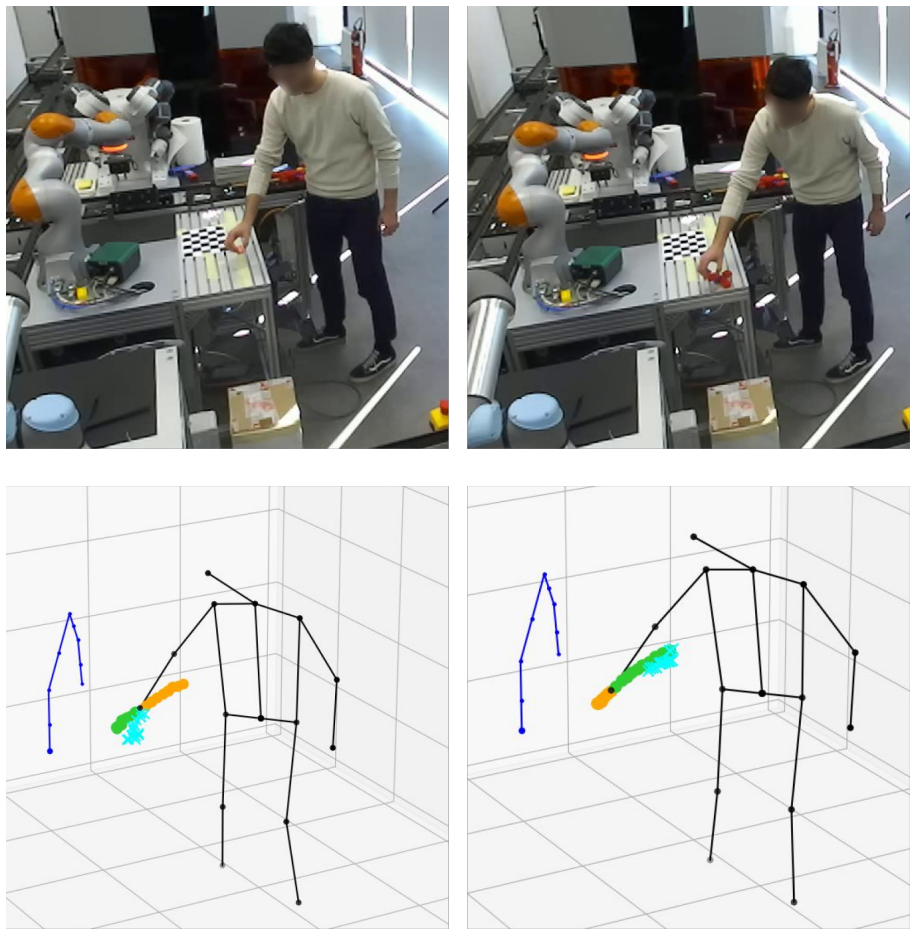} \hfill
    \includegraphics[height=6cm]{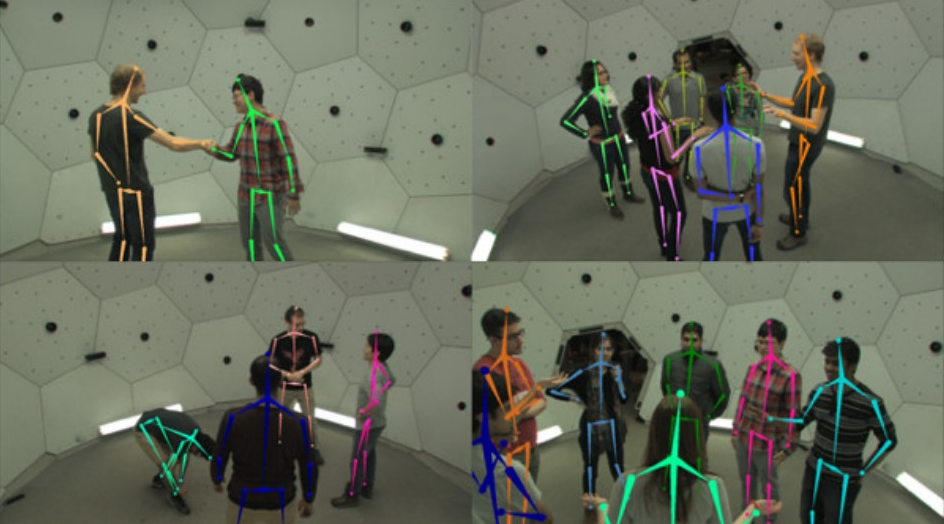}
    \caption{\textit{Top-down} and \textit{Bottom-up} approaches: an example of a \textit{top-down} apporach, the robotic arms risks to generate false positives (\emph{left}, reprinted from~\citep{chico}); the \textit{bottom-up} approach works very well with strong occlusion situation, the ideal environment excludes false positives risks (\emph{right}, reprinted from~\citep{panoptic})}
    \label{fig:2dapporaches}
\end{figure}

Pose estimation has two very closely related variants: 2D (\ie in images) and 3D (\ie in the world). Although they have quite different solution methods, very often 3D pose estimation 
is directly consequential to 2D pose estimation (mostly in multi-camera systems).

In 2D pose estimation, the goal is to accurately identifying the location of body joints in an image. 
Over the past few years, several solutions based on deep learning models have been proposed. 
Early works used convolutional filters capable of learning the physical characteristics of each joint~\citep{AE}. The use of Convolutional Neural Networks (CNN) proved to be effective in that the appearance of keypoints can be learned, ensuring robustness and speed in inference, with a remarkable ability to generalize even in the wild. The substantial difference between several CNN-based models lies in the architectural design of the networks. 

In HRNet~\citep{HRNet}, residual connections were inserted between layers having equal resolution between encoders and decoders, ensuring the flow of information regardless of upscaling and downscaling performed in the bottleneck of the network.
The most popular work based on CNNs is OpenPose~\citep{cao2017openpose}, a multi-person approach that gained a high reputation due to its ease of integration and real-time performance. EfficientPose~\citep{groos2021efficientpose} is a lighter and more accurate model version of the OpenPose that represents a strong baseline for applications usable with mobile devices.
More recent approaches have shifted the focus to the use of transformer~\citep{transformer} based models and attention-based mechanisms instead of convolutional filters.
For instance, \citep{est2Dtransformer} exploits an image token-based approach, going to segment into image sub-portions (tokens) through the use of some CNN layers, alternating with transformer blocks in order to accurately select tokens that refer to body joints.
A recent work~\citep{Zoomnet} makes use of a particular version of Higher Resolution Net (HRNet W48) to extract the body components, aggregating the basic network with hand and face estimators in order to get a complete representation of the human skeleton. Besides providing a new baseline for 2D pose estimation, not only provides information about the body, but hands and face too.

Human Pose Estimation is not only related to joints identification, rather to represent a person with skeletal model.
Thus, there are two approaches: \emph{bottom-up} and \emph{top-down}. 
The bottom-up approach consists of localizing body joints within the image, and then identify which subsets of these keypoints belong to the same person. These approaches are very useful in images with strong occlusions (\eg only the upper body) as they are able to identify joints without identifying the person first~\citep{AE,HRNet}. Nonetheless, the main disadvantage of the bottom-up approaches is the number false positives that are usually found, especially in environments with strong noise due to reflections or ``human-like'' objects (\eg robotic arms)~\citep{chico}. 
Conversely, the top-down approach involves first identifying all the individuals in the scene and then estimating keypoints for each person. Methods based on this approach provide robustness to false positives, at the cost of weak person detection in case of high body occlusions~\citep{Zoomnet}. 

In multi-view camera systems, 3D pose estimation has a simple geometric solution~\citep{hartley2003multiple}. Older works also proposed to make use of silhouettes which can be extracted reliable for a static background ~\citep{Carranza03,Balan:ICCV:2007}.

Given the position of a target point (\ie a body joint) in multiple calibrated views, it is possible to compute 3D position of the target point via triangulation.

Unfortunately this is not sufficient for developing performant 3D HPE systems. 

In fact, many errors related to system calibration and joint localization and matching concur in generating uncertainty in the 3D estimate.
To partially overcome for this effect, VoxelPose~\citep{tu2020voxelpose} introduced a triangulation network that is able to approximate the triangulation function as well as compensate for small errors in 2D estimates and calibration.

However, this generates problems at the application level: first, it easily overfits training data, and second, it is computationally expensive.
PlanSweepPose~\citep{plansweeppose} is a robust variant of VoxelPose where the 3D position of a point is estimated by minimizing the reprojection error in multiple views, while FasterVoxelPose~\citep{ye2022faster} is an improved version of VoxelPose that reduces inference time maintaining high accuracy.

Given the limited access to annotated training data with high variability, numerous models have been studied to improve generalization capabilities to new environments. 
 
RepNet~\citep{wandt2019repnet} tries to estimate the human pose with a weakly supervised approach. Starting from the idea that for every view the corresponding 3D pose is the same, the training phase forces the latent space of the model for every view to be equal, that is the actual spatial position.
More recent models like MetaPose\citep{metapose} and FLEX~\citep{flex} try to estimate 3D poses without any kind of reprojection operation and thus without the need for calibrated multi-view systems. The direction that this type of system wants to take HPE's research is increasingly toward plug-and-play systems that can generalize and be usable without the need for ground truth.

In summary, 3D pose estimators rely on robust 2D poses, but the lifting operation (\ie from 2D to 3D) introduces uncertainty which has also been addressed in recent works~\citep{WehRud2021a}. On the other hand, in many real-world applications, one can use strong priors (\eg planar motion, physical constraints) to improve reconstruction quality as in the case of gait analysis (see Sect.~\ref{sec:app:gait}).

\begin{table}[t]
    \centering
    \small
    \caption{Summary of HPE methodsd. The papers are associated by three main categories: Single-View (SV) or Multi-View (MV), RGB or RGB-D.
    }
    \label{tab:summary}
    \begin{tabular}{|l|c|c|c|c|c|}
        \hline
        \textbf{Paper} & \textbf{Approach} & \textbf{2D/3D} & \textbf{SV/MV} & \textbf{Input} & \textbf{Time} \\ \hline \hline
        ~\citep{kinectrgbd} & Random Forest  & 3D & SV & RGB-D &  \\ \hline
        ~\citep{simo2013joint} & Gen./Disc. Model & 2D\&3D & SV & RGB & \\ \hline
        DeepPose~\citep{deeppose} & CNN & 2D & SV & RGB & \\ \hline   
        Associative Embed.~\citep{AE} & Ass. Emb + CNN & 2D & SV & RGB & \\ \hline
        OpenPose~\citep{cao2017openpose} & PAF + CNNs & 2D\&3D & SV\&MV & RGB &\\ \hline
        ~\citep{tome2017lifting} & Prob. Model + Lifting & 2D\&3D & SV & RGB & \\ \hline
        ~\citep{multiviewrgbd} & Random Forest & 3D & MV & RGB-D &  \\ \hline
        VideoPose3D~\citep{videopose3d} & FCN & 3D & MV & RGB & \checkmark\\ \hline
        V2v-PoseNet~\citep{v2vposenet} & 3D CNN & 3D & SV & RGB-D &  \\ \hline      
        ~\citep{multiviewrgbd2} & CNN & 3D & MV & RGB-D &  \\ \hline
        ~\citep{wang2018robust} & Sparse Bases Repr. & 3D & SV & RGB & \checkmark \\ \hline
        Repnet~\citep{wandt2019repnet} & GAN & 3D & MV & RGB & \\ \hline
        HRnet~\citep{sun2019deep} & CNN & 2D & SV & RGB & \\ \hline
        VoxelPose~\citep{tu2020voxelpose} & 2D\&3D CNN & 3D & MV & RGB & \\ \hline
        UniPose~\citep{artacho2020unipose} & CNN+LSTM & 3D & MV & RGB & \\ \hline
        
        CanonPose~\citep{wandt2021canonpose} & MV cycle consistency & 3D & SV & RGB & \\ \hline
        ProbPose~\citep{WehRud2021a} & Normalizing Flows & 3D &SV & RGB & \\ \hline
        EfficientPose~\citep{groos2021efficientpose} & VGG + CNN & 2D & SV & RGB & \\ \hline
        PlaneSweepPose~\citep{lin2021multi} & CNN & 3D & MV & RGB & \\ \hline

        FLEX~\citep{gordon2022flex} & CNN + Attention & 3D & MV & RGB & \checkmark\\ \hline 
        MetaPose~\citep{usman2022metapose} & CNN + MLP & 3D & MV & RGB & \\ \hline
        Faster VoxelPose~\citep{ye2022faster} & 2D\&3D CNN & 3D & MV & RGB & \\ \hline
        
        AlphaPose~\citep{alphapose} & CNN \& Re-ID & 2D & SV & RGB &\\ \hline

        TCT~\citep{est2Dtransformer} & Transformer & 2D & SV & RGB & \\ \hline
        
        ~\citep{rgbdhpe1} & Depth alignment & 3D & SV & RGB-D & \checkmark \\ \hline

    \end{tabular}
\end{table}

\subsection{Marker vs. Markerless}

Marker-based MoCap are proved to be extremely accurate and robust, but still with strong limitations: \emph{(i)} they require dedicated hardware and procedures that make them expensive to buy and use, 
\emph{(ii)} retroflective markers can fall off, hamper the motions and need to be placed very accurate,
\emph{(iii)} they are often limited to closed environments with controlled conditions, and \emph{(iv)} they provide information on a limited (predetermined) number of points. Finally, trained staff is required for calibration, subject preparation, tracking and the interactive correction of mismatches.

An attempt to overcome these limitations was the GaMoCap system~\citep{biasi2015garment}, a multistereo marker based system where thousands of markers are printed on a special garment and 3D skeleton fitting is performed with structure from motion techniques.

In contrast, in HPE systems the point estimation is done by 2D estimators, which, however, 
struggle to estimate the same point between different views.
An example might be the shoulder in front and lateral view. Even for a human, it is difficult to select a pair of points that provide accurate triangulation. Automatic approaches have to optimize an N-partite graph matching problem, which is known the be NP-hard. In commercial applications, approximate solutions, \eg based on spectral clustering, are common.
Still, identifying the same point in different views is a core limitation of 3D HPE systems.

Markerless approaches, on the other side, are easier to set up and maintain (without any human intervention) and they work well on non-collaborative users, making them the most suitable technologies for clinical applications involving, for instance, infants.

The second main limitation of markerless HPE systems lies in the fact that the computational complexity for the two stages (first 2D estimation and then 3D estimation) is much higher than in MoCap systems. For this reason unlike MoCap, markerless systems often require  numerous computations, making it very difficult to run in real-time on commonly used devices such as cell phones and mid-range laptops.
Nevertheless, IoT solutions and knowledge distillation approaches have been proposed to reduce the computational costs and modern hardware and GPUs allow for very efficient inference times.

\subsection{Single-View vs. Multi-View}

\begin{figure}[t]
    \centering
    \includegraphics[height=5cm]{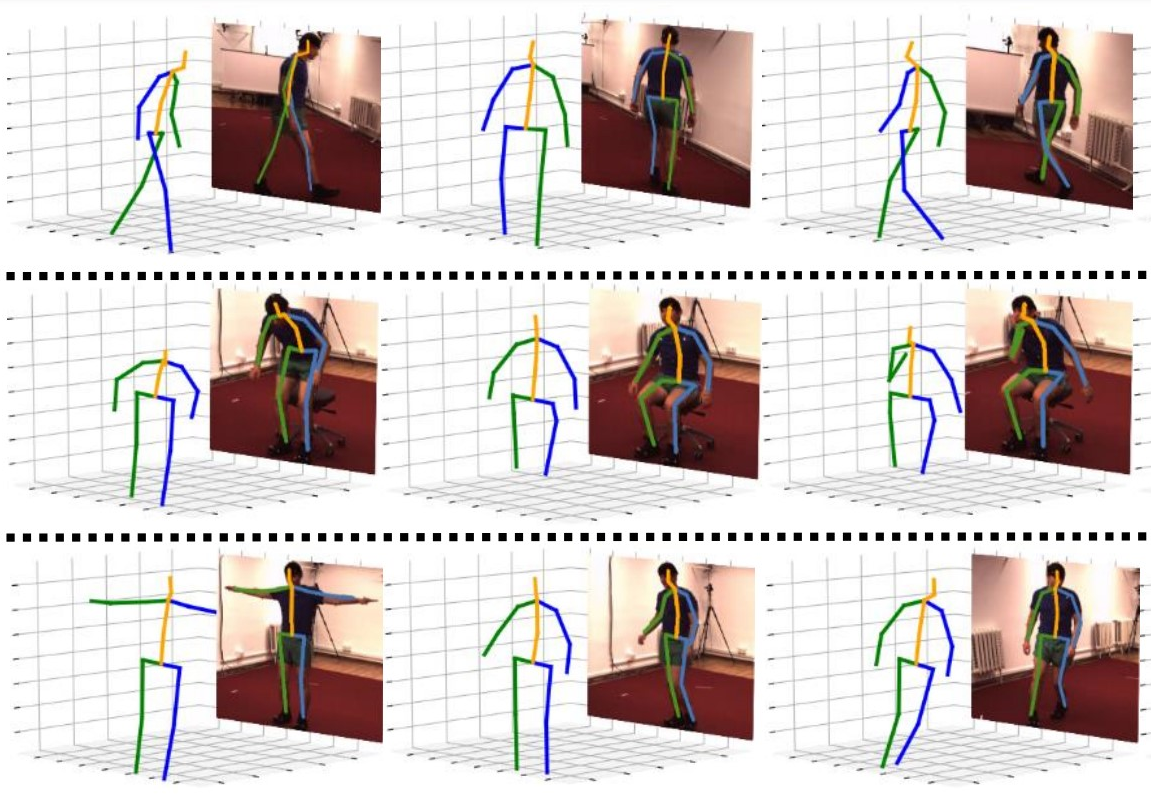} \hfill
    \includegraphics[height=5cm]{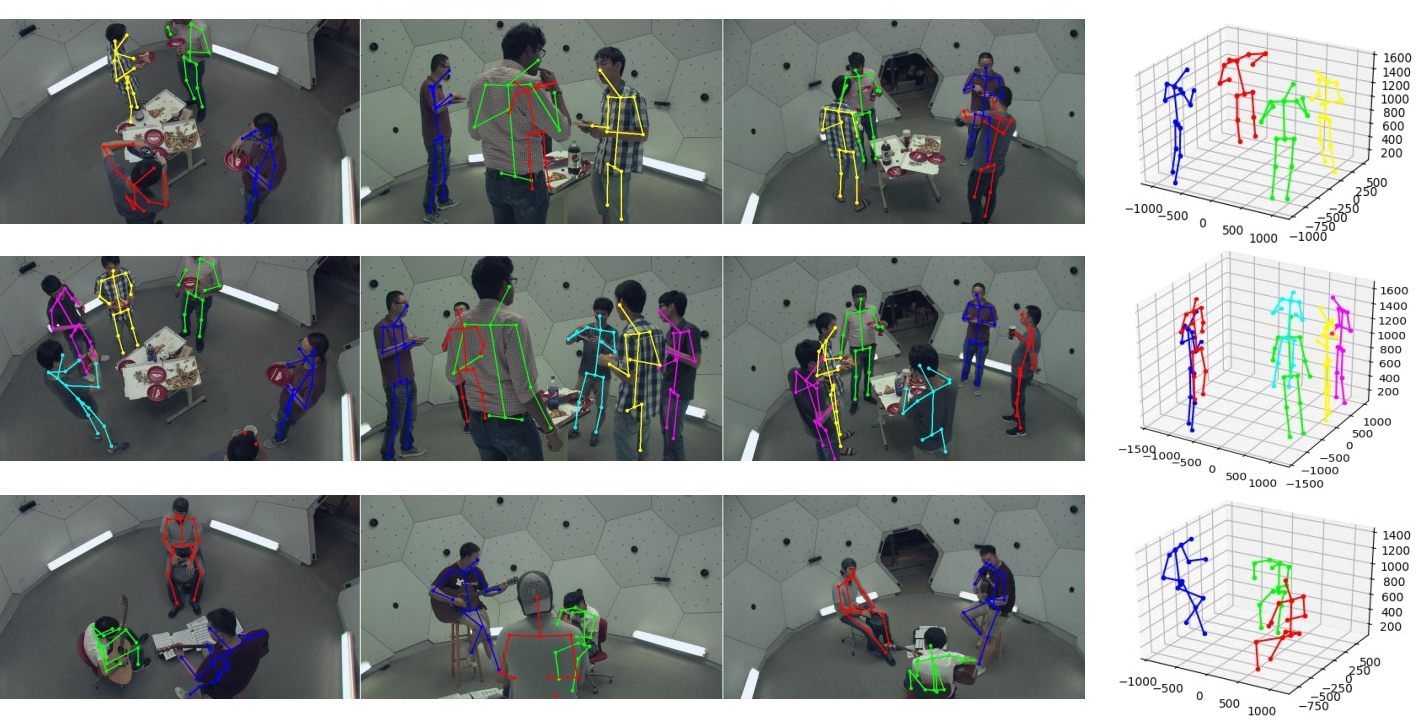}
    \caption{Examples of \emph{single-view} and \emph{multi-view} approaches. Single-view HPE on Human 3.6M (\emph{left}, reprinted from~\citep{monocular}) and an example of a multi-view HPE on CMU Panoptic (\emph{right}, reprinted from~\citep{tu2020voxelpose})}
    \label{fig:single-multi_view}
\end{figure}

The identification of a point in 3D can be traced back to a geometric problem, based on at least two different views of the same point.
Thus, the problem cannot be solved geometrically for all those setups that are formed by a single camera.
However, several works have been proposed to estimate a 3D skeleton from single view by using heuristics and priors. This operation is usually called \emph{3D lifting}.
The use of neural networks allows one to be able to estimate depth from a 2D HPE of the skeleton~\citep{deeppose,artacho2020unipose,wang2018robust,simo2013joint,tome2017lifting,wandt2021canonpose}.
The most popular technique is to use different CNN blocks to do lifting and refining the depth of the 2D pose in space. In other words, it estimates a direct mapping between the position and scale of the reprojected 2D pose with respect to the 3D pose that generated it.
At the qualitative level, monocular estimations work very well; there are very versatile frameworks such as MMPose that implements VideoPose3D~\citep{videopose3d}.
Specfically, VideoPose3D is based on temporal convolutions to provide robustness and consistency between consecutive frames, with self-supervised learning paradigms to ensure reusability of the system without the need for ground truth. In the case of monocular systems, however, this is always an estimation and not a geometric projection. For this reason, multi-camera systems are generally more accurate, and mostly designed for performing 3D human pose estimation, but they require hardware and synchronization among different cameras.Thus multi-camera setups are more expensive compared to monocular HPE systems, which can hold very low costs and be more usable in everyday life. In figure \ref{fig:single-multi_view} an examples of different setups, showing the complexity of a multiview setup compared to just a monocular one. 

\subsection{RGB vs. RGBD}

By definition, images are a projection of the 3D environment into 2D, mixing up a third dimension. 
Hence, systems that rely on RGB only, often struggle to infer the depth and try to get insights of it in some way, directly or indirectly.
An alternative technology are depth cameras, either based on structured light or time-of-flight. More common nowadays are devices integrating both channels (RGB and depth) to produce so called RGB-D images.
Unlike traditional RGB cameras, RGB-D devices output two distinct images, one color image and a depth map where each pixel returns the distance between the camera lens and the corresponding point in the 3D world. Since sensors often have different frame rate, sensing ranges and resolution, interpolation is often performed to align the RGB-D sensors for follow up tasks. This can also induce inaccuracies into the overall system.
Similar results can be obtained, with less accuracy, using (small baseline) stereocameras and disparity map computation.

The most popular cameras on the market include Zed, Microsoft Kinect, and RealSense. 
Using RGB-D cameras it is relatively easy to combine 2D estimator with depth maps to derive 3D poses.
Early approaches to pose estimation using RGB-D cameras exploit splitting the problem in two tasks~\citep{kinectrgbd}: a body keypoint localization is made for each pixel in the image and then the final joint position is inferred. 
Similarly, other attempts have been made by analyzing probabilistic inference by optimizing the pose of kinematic models~\citep{zhu}.
More recent techniques based on CNN employ two different methodologies: the first~\citep{v2vposenet} consists of using 3D Convolutions to identify the joint in the three-dimensional space provided by the acquired depth.
The second method~\citep{rgbdhpe1} on the other hand aims to estimate the skeleton using a 2D estimator in the image, moving on to retrieve the spatial information from the depth camera to estimate the distance.
The accuracy that can be achieved with systems of this type, however, turns out to be lower than the performance of a multi-view system.
This performance gap is mostly due to self-occlusions.
Especially in systems such as~\citep{v2vposenet,rgbdhpe1}, if the arms are estimated correctly behind the torso the detected distance will be the same, resulting in a flat pose.
This problem, however, has been solved by systems that make use of multi-view RGB-D cameras~\citep{multiviewrgbd,multiviewrgbd2}, where 3D pose is computed for each view, then skeletons are aligned and recorded for increased spatial rigor and consistency.

\subsection{Temporal reasoning}

Until now, the problem of pose estimation has been discussed only referring to frames that were intended atomically.
However, when we are faced with applying HPE algorithms to sequences of frames, we often encounter problems in that the slightest variations in pixels from frame to frame can cause chattering.
To solve this problem, temporal filtering methodologies, such as Kalman filters~\citep{rgbdhpe1}, have been incorporated within the models. 
Other methodologies such as VideoPose3D ~\citep{videopose3d} make use of 3D lifting convolutions that process poses sequentially, providing results with temporal consistency.
The use of filters to make the pose sequences smooth tend to reduce what is the range of motion of the person, affecting the spatial coherence of the system.
In addition, the use of effective filters is not possible in the case of real-time models without introducing latency in the system. 

\section{Biomedical applications of HPE}
\label{sec:applications}

It is widely recognized that there is a need for new technologies for human motion analysis. Current methods are either subjective (\eg clinical scales), expensive, potentially inaccessible to most clinicians (\eg MoCap systems), or provide only limited information about specific, predefined features of movement~\citep{cherry2023opportunities}.
HPE offers clear and significant potential for applications in biomedical field, as it enables quantitative measurement of human movement kinematics in virtually any setting with minimal cost, time investment, and technological requirements.

Based on an accurate analysis of recent literature in the field, we identified four main areas where HPE can be effectively applied as an assistive tool for clinicians or an automatic support for patients. In the following, we will analyse each field, summarizing recent works and discussing potentialities of HPE to further expand.

\subsection{Motor Development Assessment}

Motor development is a process aimed at achieving full mobility and independence by an infant. In healthy developing infants, the sequence of movements is the same and repetitive.
The detection of abnormalities in an infant’s motor development is crucial for the early diagnosis of neurological disorders, thus allowing to start of the therapeutic process,  reducing the likelihood of sensory disorders, coordination problems, and postural problems. 

\begin{figure}[t]
    \centering
    \includegraphics[height=6cm]{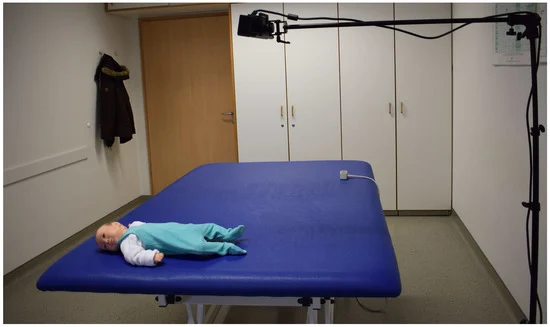} \hfill
    \includegraphics[height=6cm]{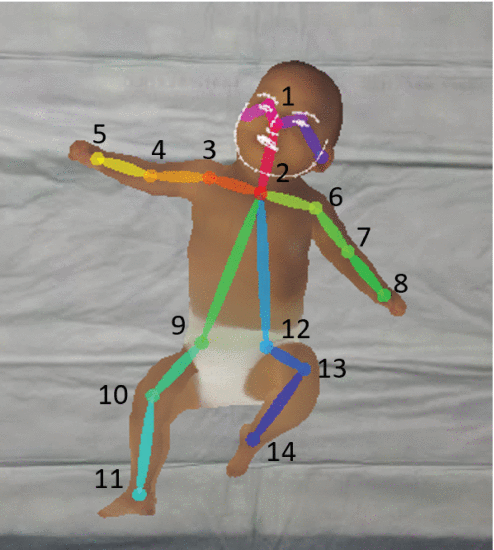}

    \caption{Typical acquisition setup for General Movement Assessment (GMA) on newborns and infants with the subject lying on a uniform coloured bed and a camera above it in perpendicular direction (\emph{left}, reprinted from~\citep{khan2018detection}), and an example of skeletal reconstruction of an infant with a simplified 14-joints model (\emph{right}, reprinted from~\citep{mccay2020abnormal}).}
    \label{fig:mda_mccay}
\end{figure}

Current video-based approaches suffer from the time-intensive but necessary process of manually coding child behaviors of interest by clinicians. Pose estimation technologies offer the opportunity to accelerate video coding to capture specific behaviors of interest in such developmental investigations. Such approaches could further help offering a cost-effective and scalable way to distinguish between typical and atypical development~\citep{stenum2021applications}.

The main issue in the adoption of automatic HPE models in motor development assessment is related to the age of the subjects, typically in the range 0 to 4 years. Kids have different anatomical proportions with respect to adults, which represent the majority of training data for HPE systems.
A challenging task in this context is to transfer knowledge acquired from large datasets of adult people to infants by using only a limited amount of task specific data. 
For this reason, the majority of studies involving newborns (0 to 2 months of age) and infants (2 to 12 months of age) exploit traditional geometry based approaches, while for toddlers (1 to 4 years of age) we witness an effort in transferring knowledge of deep learning models.

\begin{table}[t]
    \centering
    \caption{Summary of papers addressing motor development assessment with HPE techniques (in chronological order). Target: N=newborn, I=infant, T=toddler.} 
    \label{tab:mda}
    \small
    \begin{tabular}{|l|c|c|c|c|c|}
        \hline
        \textbf{Study} & \textbf{Input} & \textbf{Estimate} & \textbf{HPE} & \textbf{Classification} & \textbf{Target} \\ \hline \hline
        \citep{hesse2017body} & depth & 3D & Random Ferns + PCA & -- & N \\ \hline
        \citep{khan2018detection} & RGB & 2D & DPM + SVM & -- & N/I \\ \hline
        \citep{hesse2019learning} & RGB-D & 3D & SMIL (from SMPL) & -- & N \\ \hline
        \citep{ihlen2019machine} & RGB & 2D & Optical Flow & LDA & I \\ \hline
        \citep{moccia2019preterm} & depth & 2D & CNNs & -- & N \\ \hline
        \citep{chambers2020computer} & RGB & 2D & OpenPose & NGBS & I \\ \hline
        \citep{doroniewicz2020writhing} & RGB & 2D & OpenPose & SVM, RF, LDA & N \\ \hline
        \citep{li2020three} & RGB-D & 3D & OpenPose & -- & I \\ \hline
        \citep{mccay2020abnormal} & RGB & 2D & OpenPose & ANN & N \\ \hline
        \citep{ni2020siamparsenet} & RGB & 2D & SiamNet & -- & I \\ \hline
        \citep{adde2021motion} & RGB & 2D & CNN & -- & I \\ \hline
        \citep{carbonari2021end} & depth & 2D & Mask R-CNN & -- & I\\ \hline
        \citep{reich2021novel} & RGB & 2D & OpenPose & ANN & I \\ \hline
        \citep{sakkos2021identification} & RGB & 2D & OpenPose & ANN & I \\ \hline
        \citep{groos2022development} & RGB & 2D & EfficientPose & GCN & I/T \\ \hline
        \citep{gong2022preterm} & RGB & 2D & HRnet & CTR-GCN & I \\ \hline
        \citep{ni2023semi} & RGB & 2D & SiamNet + GAN & -- & I \\ \hline

    \end{tabular}
\end{table}

Early works focused on fitting a skeletal model starting from body joints localization or body parts segmentation. The testing setup is common for all these studies, \ie a Microsoft Kinect camera positioned on top of a testing bed to acquire newborns motion. 
Depth channel is used in~\citep{hesse2017body} to extract Random Ferns features trained on synthetic data and process them to localize body joints. 
RGB channel is employed in~\citep{khan2018detection} instead, where Deformable Part Models (DPM) are used to segment body parts and then a structured SVM classifier assigns labels to each segment. 
Multimodal data analysis is introduced in~\citep{hesse2019learning}, where for the first time both RGB and depth channels are exploited to directly fit a 3D virtual model of a baby generated as an extension of SMPL~\citep{loper2015smpl} model to infants.
All these works are designed as assistive tools for clinicians: they provide a visualization and quantitative reconstruction of the skeletal model of the baby. Some works also provide indication of the joint angles. Anyway, a common point is that they do not aim at recognizing atypical subjects or providing measures on how the observation differs from typical data.
A step forward is presented in~\citep{ihlen2019machine} that performs body parts segmentation based on optical flow and particle matching; then motion features are extracted and an LDA model is trained to predict high/low risk of cerebral palsy in infants.
While early work rely on relatively simple traditional computer vision methods for pose estimation, more accurate HPE approaches can boost performances.

OpenPose~\citep{cao2017openpose} library has been proved as a solid asset for estimating the risk of neurological disorders in newborns and infants.
\citep{doroniewicz2020writhing} attempts to automatically detect writhing movements on newborns within three days of life. Movement features are automatically extracted from baby poses and used to distinguish between newborns with good level of writhing movements against those showing a poor repertoir, \ie a lower quality of movement in relation to the norm.
In~\citep{mccay2020abnormal} reconstructed poses are used to extract normalised pose-based feature sets, namely Histograms of Joint Orientation 2D (HOJO2D) and Histograms of Joint Displacement 2D (HOJD2D). These features are then fed to five different deep learning models was aim is to predict if observed movements are indicative of typically developing infants (Normal) or that may be of concern to clinicians (Abnormal).
A Na\"{i}ve Gaussian Bayesian Surprise (NGBS) is applied in~\citep{chambers2020computer} to calculate how much infant’s movements deviate from a group of typically developing infants and output an indicator of risk for atypical motor development.
In~\citep{li2020three} 3D pose estimation is achieved starting from 2D body poses by using the mapping function between pixels in the RGB and depth channels of the RGB-D scan with a Kinect sensor.
All previous works use a simplified OpenPose model with 14 body joints. A more complete model with 25 body joints is used in~\citep{reich2021novel} instead. Poses represent input vectors for a shallow multilayer neural network to discriminate fidgety from non-fidgety movements in infants.
More sophisticated models like EfficientPose~\citep{groos2021efficientpose} are also applied to MDA in combination with Graph Neural Networks to predict cerebral palsy in infants at high risk~\citep{groos2022development}.

In the last four of years, there has been an increased interest in developing dedicated deep learning based approaches for baby pose estimation.
In~\citep{moccia2019preterm}, limb-pose estimation is performed using a deep-learning framework consisting of a detection and a regression convolutional neural networks (CNN) for rough and precise joint localization, respectively. The CNNs are implemented to encode connectivity in the temporal direction through 3D convolution.
As an extension of this work, \citep{carbonari2021end} proposes an end-to-end learnable convolutional neural network architecture based on Mask R-CNN model for limb-pose estimation in NICUs facilities.
More recetly, a new infant pose estimation method has been proposed in~\citep{wu2022supine}. The main contribution is represented by a novel keypoint encoding method, called \emph{joint feature coding} (JFC), to produce high-resolution heatmaps with a CNN model.
The system presented in~\citep{gong2022preterm} first extracts trajectories of the body joints with HRNet~\citep{HRNet} to learn the representations that are most relevant to the movement while avoiding the irrelevant appearance-related features, and then returns a normal/abnormal movement binary classification with CTR-GCN~\citep{chen2021channel}. 
Finally, to overcome the limited availability of labelled training data in infant pose estimation, a semi-supervised body parsing model, called \emph{SiamParseNet} (SPN), to jointly learn single frame body parsing and label propagation throughout time is proposed in~\citep{ni2020siamparsenet}. SPN consists of a shared feature encoder, followed by two separate branches: one for intra-frame body parts segmentation, and one for inter-frame label propagation. In~\citep{ni2023semi} the same authors complement SPN with a training data augmentation module, named Factorized Video Generative Adversarial Network (FVGAN), to synthesize novel labeled frames by decoupling foreground and background generation.

In line with the general trends of digital medicine, some recent works tried to simplify the acquisition setup, opening to remote diagnosis. In~\citep{adde2021motion} videos of infants were taken at home using handheld smartphones and the In-Motion-App; a simplified skeletal model with only 7 joints (head, chest, pelvis, wrists, and ankles) is fit and tracked by a convolutional neural network trained on about 15k video frames of high-risk infants. Fidgety movements were then classified as continuous (FM++), intermittent (FM+), sporadic (FM+/-), abnormal (Fa) or absent (FM-), according to Prechtl's method~\citep{ferrari2004prechtl}. This was the first automatic system tested on video recordings from handheld smartphones.

Despite being proved to perform as well as an experienced clinician in identifying babies with high risk of neurological disorders~\citep{groos2022towards}, the adoption of such systems in the clinical practice is slow because of the lack in interpretability of results, \ie doctors can hardly understand the reasons why a machine makes some predictions in case of disagreement.
A big step forward in this sense is represented by~\citep{sakkos2021identification}. This work introduces a framework that provides both prediction and visualization of those movements that are relevant for predicting a high-risk score. First the 2D skeletal pose is detected on a per-frame basis using OpenPose, hence, for each pose body parts are segmented and processed by a specific branch to learn a part-specific spatiotemporal representation using an LSTM. Finally, the output is passed to a shallow fully connected network for classification as well as to a visualization tool.

\subsection{Neuromuscolar Rehabilitation}

Human motion analysis for rehabilitation has been an active research topic for more then 30 years now.
Markerless motion capture technologies offer clear and significant potential for applications in rehabilitation, as they enable quantitative measurement of human movement kinematics in virtually any setting
with minimal cost, time investment, and technological requirements~\citep{cherry2023opportunities}.
Such systems represent valuable tools for many use cases. 
In clinical work, accurate reconstruction allows physiotherapists to monitor the progress of patients in their rehabilitation journey.
In telerehabilitation setups, patients can perform their workout autonomously with supervision of an AI.
Accurate estimates of skeletal models are also necessary for a safe and effective implementation of robotic assisted rehabilitation protocols.

A supporting system for managing therapeutic exercises should consist of two central tools: capturing human motion in real-time, and quantitatively evaluate the performance.
In order to be effective an HPE based platform used in rehabilitation must ensure correct exercise performance of the client and automatic detection of wrong executing.

Assessment studies like~\citep{sarsfield2019clinical} observed that current (at that time) markerless HPE systems were mostly inadequate for correctly assessing rehabilitation exercises due to poor accuracy and inability to handle self occlusions.
Nevertheless, newer approaches were presented in the last years and researchers proposed variations of standard HPE models that are specifically designed to perform well in rehabilitation scenarios.
We should also consider that in this context patients could have appearance significantly different from the typical distributions used in publicly available annotated training datasets. Consider the fact that patients are usually recovering from surgeries where arts could be amputee, or they could be following therapies for obesity related postural problems. In these cases, pre-trained HPE models can easily fail in tracking joints or fitting the skeletal model. For this reason annotated datasets have been provided for the specific domain.

In~\citep{xu2022multiview} a framework for accurate 3-D pose estimation using multiview videos is presented. The pipeline is divided in two steps: First, a Stacked Hourglass Network~\citep{newell2016stacked} associated with a coarse-to-fine heatmap shrinking (CFHS) strategy generates localizes joints on each image; Second, a spatial-temporal perception network fuses 2D results from multiple views and multiple moments (\ie leverages temporal information) to generate 3D pose estimation.
A combination of RGB and depth channels captured by a Microsoft Kinect device are used in~\citep{wu2020human}. 2D poses are extracted from RGB images using Part Affinity Fields~\citep{cao2017realtime} and then exploit mapping between the two channels to estimate 3D body joints locations. 
Such models perform very well in their specific contexts, but it is also computationally demanding, making it only suitable for usage in clinic.

For telerehabilitation scenarios instead, effort was made to design a neural network model that is both accurate and lightweight in computation.
\citep{li2020human} adopts HRNet as the backbone and leverage the modules in MobileNetv1\citep{howard2017mobilenets} to simplify the original network. Meanwhile they also leverage dilated convolution~\citep{mehta2019espnetv2} for its spatial features extraction capability. The resulting network, called Extremely Efficient Spatial Pyramid (EESD), is further enriched with an attention module and can run smoothly on a mid level smartphone. 
A web based application called NeuroPose is presented in~\citep{rick2019neuropose}. The backbone of pose estimation is PoseNet~\citep{v2vposenet} model deployed using tensorflow.js framework, a browser based implementation that uses common libraries like WebGL to access local GPU resources when available, and otherwise can scale back parameters to support a wider range of hardware.
Alternative solutions are represented by the two opposite directions of cloud and edge computing. 
In~\citep{prima2019single} 2D poses are provided by OpenPose library, 3D lifting exploits heuristics and prior information, and joint angles are computed from skeleton model. Here the IoT paradigm is exploited to allow the application to work on a smartphone, while computationally expensive parts are running on a cloud service.
Conversely, \citep{dos2021deeprehab} introduces DeepRehab, a bottom up model inspired by PoseNet that estimates joints' positions with a fully convolutional architecture with ResNet101 as a backbone for feature extraction. The code runs on an Edge TPU device that is dedicated to this computation, while the patient can interact with a smartphone app.

A new trend is represented by robotic assisted rehabilitation, particularly useful for patients with motor disorders caused by stroke or spinal cord disease since rehab robots can deliver high-dosage and high-intensity training.
In this case the focus is on guaranteeing safety during execution of exercises.
A comparison of two popular HPE models, namely OpenPose and Detectron2, for angles estimation in upper body joints is provided in~\citep{hernandez2021human}. This study concludes that OpenPose has accuracy compared with expert therapists and a frame rate of about 10fps, making it suitable for real world applications.
Thus, OpenPose is used in~\citep{tao2020trajectory} to track human poses when executing an exercise supported by an expert; then a robot plans trajectories to substitute the physiotherapist in assisting patients while replicating the same movement. Again, 3D poses are obtained from depth channel.

\begin{table}[t]
    \centering
    \caption{Summary of papers addressing neuromuscolar rehabilitation with HPE techniques (in chronological order).} 
    \label{tab:rehab}
    \small
    \begin{tabular}{|l|c|c|c|c|}
        \hline
        \textbf{Study} & \textbf{Input} & \textbf{Estimate} & \textbf{HPE} & \textbf{Target} \\ \hline \hline
        \citep{prima2019single} & RGB & 3D & OpenPose & full \\ \hline
        \citep{rick2019neuropose} & RGB & 3D & PoseNet & full \\ \hline
        \citep{li2020human} & RGB & 3D & HRnet & lower \\ \hline
        \citep{tao2020trajectory} & RGB-D & 3D & OpenPose & full \\ \hline
        \citep{wu2020human} & RGB-D & 3D & PAF & full \\ \hline
        \citep{dos2021deeprehab} & RGB & 2D & DeepRehab & lower \\ \hline
        \citep{xu2022multiview} & RGB & 3D & SHN + CFHS + EVLT & full \\ \hline

    \end{tabular}
\end{table}

\subsection{Gait and Posture Analysis}
\label{sec:app:gait}

Gait analysis is the systematic study of body movements that are responsible for locomotion in human beings, \ie walk, run, jump, \etc 
It is widely used to identify gait abnormalities and to propose appropriate treatments. It is an essential functional evaluation in clinical applications for recognizing health problems related to posture and movement and monitoring patients’ status. 
Succesfull applications of HPE in this field are mostly employing either OpenPose\citep{cao2017openpose} or DeepLabCut~\citep{mathis2018deeplabcut}, while a wider comparison among different HPE methods is provided in~\citep{washabaugh2022comparing} and \citep{kanko2021concurrent}.

Gait is most often a planar task, \ie patients are asked to walk on a straight line, two planes are of particular interest in these analysis: sagittal and coronal. Thus, two cameras are needed to accurately monitor 3D poses: one placed in front of the user and one laterally. Moreover, the goal of gait analysis is not to accuratly monitor the whole body, rather to extract quantitative evaluation of specific movement features, \ie statistics on joint angles.
Several studies focused on the lateral view since it can provide more relevant information in many scenarios. Three studies have examined 2D monocular applications of DeepLabCut against manual labelling or marker-based methods in sagittal view in different conditions: underwater running~\citep{cronin2019markerless}, countermovement jumping~\citep{drazan2021moving}, and walking in stroke survivors~\citep{moro2020markerless}. These studies consistently conclude that, despite having a higher average error with respect to marker based approaches, markerless HPE can be effectively used in gait and posture analysis.
Similar results are obtained by~\citep{viswakumar2022development} using OpenPose to estimate knee angles. 

The main issue with single camera setups is represented by self-occlusions.
To overcome this problem, \citep{serrancoli2020marker} and \citep{stenum2021two} used two cameras placed on either side of the sagittal plane, each one only monitoring the closest limbs and ignoring the rest. Nevertheless, both studies showed poor results in terms of accuracy and robustness, requiring frequent intervention of a user.

Temporal information has been examined in two separate studies using 2D monocular HPE combined with projection mapping~\citep{shin2021quantitative} and with 3D musculoskeletal model~\citep{azhand2021algorithm} to subsequently extract gait parameters measures, finding strong correlations when compared to the GAITRite pressure walkway\footnote{\url{https://www.gaitrite.com/}}. 
An end-to-end model is presented in~\citep{sokolova2019pose}. Starting from 2D pose estimation from OpenPose, authors crop patches around the joints and compute optical flow in these regions; finally a convolutional neural network is trained to directly output gait descriptors.

Gait analysis in biomedical applications often requires the computation of the inverse dynamics. It implies methods for computing forces and the torques based on 3D motion of a body and the body's inertial properties (e.g. masses of each limb). It usually requires additional information, e.g. ground reaction forces, stemming from force plates. 
Brubaker et al. ~\citep{Brubaker07,BrubakerICCV2009} use an articulated body model to infer joint torques and
contact dynamics from motion data. They accelerate the optimization procedure by introducing additional root forces and effectively decoupling the problem at different time frames. 
Researchers also proposed to learn a direct mapping from a motion parametrization (joint angles) to the acting
forces (joint torques) on the basis of MoCap data: Johnson and Ballard ~\citep{Johnson_Ballard_2014}
investigate sparse coding for inverse dynamics regression and Zell et.al. ~\citep{Zell09} introduced a two dimensional statistical model for human gait analysis and extended it later to a 3D model and a neural network based inference ~\citep{ZelRos2020}
While most approaches solely focus on walking motions, further works
~\citep{Vondrak08,Duff11} consider a wider range of motion types.
A physical driven prior is also suitable to support statistical approaches e.g. Wei et.al.  ~\citep{wei11} 
 use a maximum a posteriori approach to synthesize a wide range
of physically realistic motions and motion interactions.
Many human pose estimation methods
further rely either on anthropometric constraints ~\citep{akhter2015pose,Wang2014RobustEO} to stabilize the computations.

As Parkinson's disease represents a major field of investigation with gait analysis, several works focused on these class of patients.
For instance, \citep{shin2021quantitative} processed monocular frontal videos for providing temporospatial outcome measures (\ie step length, walking velocity and turning time). \citep{martinez2018accuracy} used OpenPose to examine walking cadence and automate calculation of an anomaly score for monitoring progression of the disease. \citep{connie2022pose} uses a two cameras setup (frontal and lateral) and extracts eight spatio-temporal features from 2D trajectories of body joints estimated with AlphaPose~\citep{alphapose}. These features are then used in a random forest classifier to distinguish between classes of disease severity. Similarly, \citep{sato2019quantifying} proposed an unsupervised approach to quantify gait features and extract cadence from gait video by applying OpenPose.

\section{Commercial systems}
\label{sec:commercial}
Several solutions based on HPE systems have been developed commercially for the analysis of person movement and pose.
The most popular \citep{contemplas,theiamarkerless,indatalabs,captury} are systems all based on markerless technology. Of these systems, most are purely designed to be able to do motion analysis in the medical/sports field.
All of the proposed systems make use of multiview models, with the exception of \citep{indatalabs} which also allow pose estimation in single view.
In addition, \citep{indatalabs} and \citep{captury} provide software capable of using non-proprietary camera setups, allowing a wide range of setup choices based on end-user needs.
Each solution offers dedicated frameworks and suites with integrations with Unity and Blender, so that motion can be obtained and analyzed not only based on the skeleton but also with body mesh.
Actual performance in terms of error is not reported by the site developers. 

\section{Discussion}
\label{sec:discussion}

\subsection{Limitations of HPE in Biomedicine} 

Clinical application of markerless HPE is precluded by some technical and practical limitations.
First, there is a need for extensive evaluation studies like~\citep{carrasco2022evaluation} that will allow to better understand how well these methods, designed and trained on healthy population, can be trasferred to clinical population with antropometric models that can significantly differ from average adults, \eg infants or obese people. 
Hence, there is a need of accessible annotated data for retraining models on these typologies of subjects that are of particular interests in medical applications.
Second, there is a need for accessible software that can be used out-of-the-box from non technical users with minimal effort. This is witnessed by the fact that, despite being no more state-of-the-art, OpenPose is still the most used approach in medical field.
Third, despite being low demanding in terms of acquisition setups, markerless HPE systems are relatively expensive in terms of computational power, raising barriers to the deployment on mobile devices.

\subsection{Opportunities}

All the limitations listed above, are actually opening new opportunities for researchers and technicians.
Indeed, while for HPE in-the-wild the generalization capability of a method is an important feature, for many medical applications it is actually not so relevant. Considering the case of a patient that needs for a long rehabilitation program, spending a few sessions to overfit the model to the specific user can represent an acceptable cost in the face of improved accuracy. A study in this direction is the patient-specific approach of~\citep{chen2018patient}, and we argue that other strategies like continual (\ie incremental) learning or domain adaptation can be effectively applied in this context.
As for the lack of annotated data, there is a range of studies that try to transfer learning from synthetic data to real-world data. While this is not trivial, it would allow for generating data with high variability, as well as controlling characteristics of the specific subject categories. Data can be effectively generated with Generative Adversarial Networks and used to fine-tune pre-trained general-purpose models. 
Finally, we believe that computational issues will be soon overcome by technological advancement. Nevertheless, recent studies developed solutions that can leverage either computing on dedicated edge devices, using cloud computing under the paradigm of Internet-of-Things, or mixing the two as proposed, for a different scenario, in~\citep{cunico2022split,capogrosso2023split}.

\section{Conclusions}

In this paper, we reviewed the state of the art in markerless Human Pose Estimation with applications to the biomedical field.
First, we examined the main features of HPE approaches generating a taxonomy that allows to map proposed methods and discriminate those approaches that are not suitable for specific applications.
Then we identified three areas of medicine and healthcare where HPE is already in use (or under investigation): namely motor development analysis, neuromuscular rehabilitation and gait \& posture analysis. For each of these fields we presented an overview of the state-of-the-art considering peculiarities and trends followed by researchers and practitioners.
In our analysis, we included 25 different approaches for Human Pose Estimation and more than 40 studies of HPE applied to the three areas identified above.
We can draw the conclusion that markerless HPE offers great potential for extending diagnosis and rehabilitation outside hospitals and clinics, towards the paradigm of \emph{remote medical care}. Current approaches are limited in accuracy with respect to the gold standard in clinical practice, \ie marker-based motion capture systems, still, they can complement video analysis with lots of useful information that allow clinicians to make more informed decisions, saving a huge amount of time.

\bibliographystyle{Frontiers-Harvard} 
\bibliography{hperefs}

\end{document}